\documentclass[sigconf]{acmart}
\usepackage{multirow}
\usepackage{enumitem}
\AtBeginDocument{%
  \providecommand\BibTeX{{%
    \normalfont B\kern-0.5em{\scshape i\kern-0.25em b}\kern-0.8em\TeX}}}

\begin{document}

\title{Programming Language Agnostic Mining of Code and Language Pairs with Sequence Labeling Based Question Answering}






\author{Changran Hu$^1$, Akshara Reddi Methukupalli$^2$, Yutong Zhou$^2$, Chen Wu$^3$, Yubo Chen$^4$}
\affiliation{
  \institution{$^1$University of California, Berkeley \country{USA}\\ $^2$Tencent Cloud, Shanghai, China\\ $^3$Department of Electronic Engineering, Tsinghua University, Beijing \country{China}}
}
\email{{changran_hu,akshara_methukupalli,yutong_zhou}@berkeley.edu,chewu@tencent.com,ybch14@gmail.com}


\begin{abstract}

Mining aligned natural language (NL) and programming language (PL) pairs is a critical task to NL-PL understanding.
Existing methods applied specialized hand-crafted features or separately-trained models for each PL.
However, they usually suffered from low transferability across multiple PLs, especially for niche PLs with less annotated data.
Fortunately, a Stack Overflow answer post is essentially a sequence of text and code blocks and its global textual context can provide PL-agnostic supplementary information.
In this paper, we propose a Sequence Labeling based Question Answering (SLQA) method to mine NL-PL pairs in a PL-agnostic manner.
In particular, we propose to apply the BIO tagging scheme instead of the conventional binary scheme to mine the code solutions which are often composed of multiple blocks of a post.
Experiments on current single-PL single-block benchmarks and a manually-labeled cross-PL multi-block benchmark prove the effectiveness and transferability of SLQA.
We further present a parallel NL-PL corpus named Lang2Code automatically mined with SLQA, which contains about \textbf{1.4M} pairs on \textbf{6 PLs}.
Under statistical analysis and downstream evaluation, we demonstrate that Lang2Code is a large-scale high-quality data resource for further NL-PL research. 

\end{abstract}
\maketitle

\section{Introduction}
Recent years, a new suite of developer assistance tools based on natural language processing (NLP) have started to burgeon, such as code retrieval~\cite{code_retrieval_miltos,code_retrieval_yi, csn}, code summarization~\cite{code_summarization_miltiadis, codenn} and code synthesis~\cite{code_synthesis_nicholas, code_synthesis_desai, code_synthesis_quirk, code_synthesis_yin, pymt5}.
Since data-driven methods have shown great promise on these tasks~\cite{conala}, high-quality parallel natural language (NL) and programming language (PL) corpora have gradually become essential resources, making NL-PL pair mining a critical task for these downstream applications.

Existing methods acquired the aligned NL-PL pairs through handcrafting specialized features~\cite{codenn, conala} or training separate models~\cite{staqc} for each PL.
For example, \citeauthor{conala}~\shortcite{conala} proposed to combine PL-specific handcrafted rules with neural network models.
\citeauthor{staqc}~\shortcite{staqc} proposed to incorporate the local textual context around code blocks and train separate models for Python and SQL data.
However, these methods are highly dependent on large amounts of single-PL training data for each PL, making it difficult to directly migrate them to niche PLs that lack training data.

\begin{figure}[t]
\centering
\includegraphics[width=.48\textwidth]{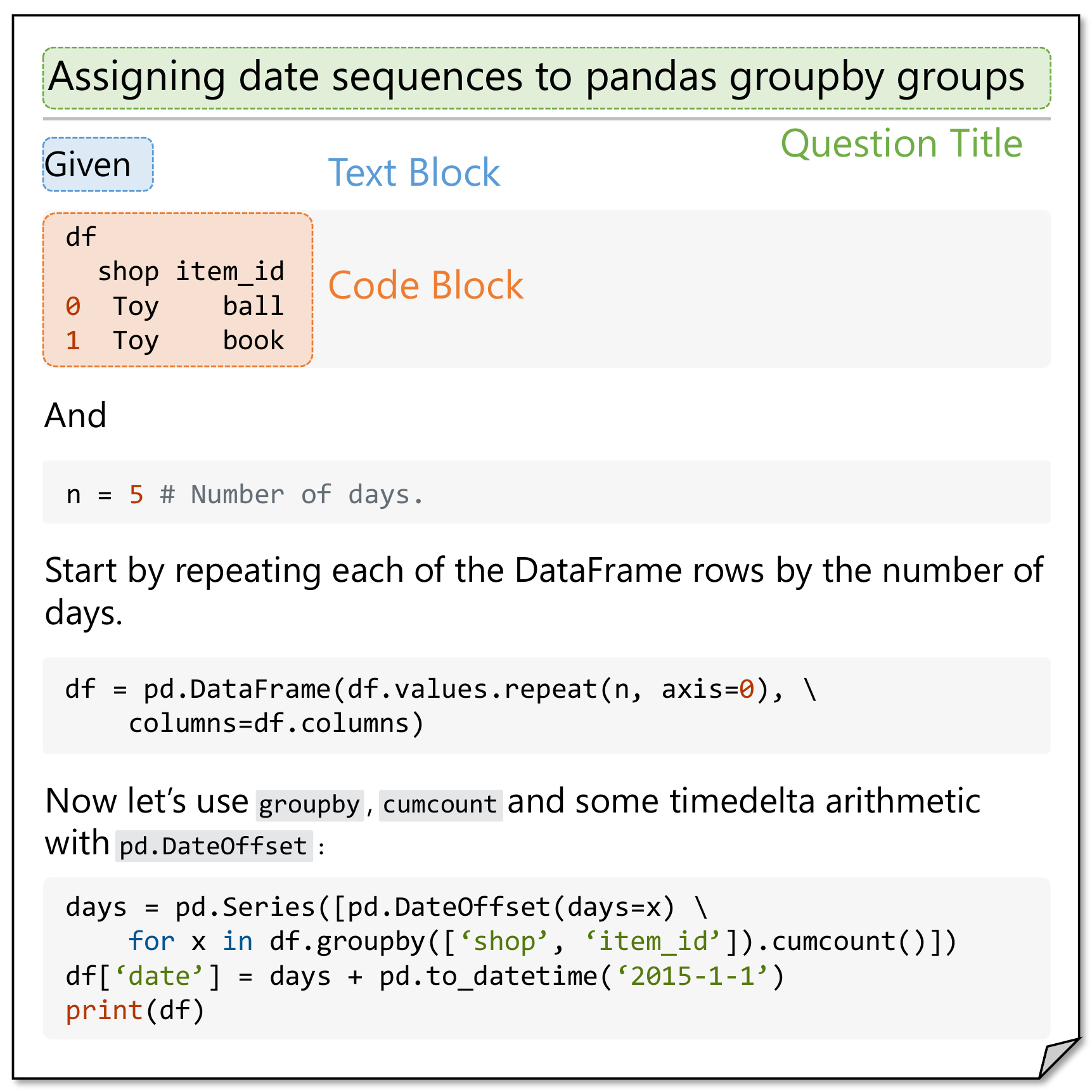}
\caption{An example of a Stack Overflow answer post, which contains a NL question title and a sequence of text and code blocks. The solution of this post consists of the last two code blocks.}
\label{fig:intro_example}
\end{figure}

Fortunately, we observe that a Stack Overflow (SO) answer post is essentially a sequence of text and code blocks.
The global textual context of the post can provide PL-agnostic supplementary information for mining correct code blocks corresponding to the post's NL title.
For example, in Figure \ref{fig:intro_example}, even if we remove the content of the code blocks, the remaining PL-agnostic context ``\textit{Given $\mathbb{C}_1$ and $\mathbb{C}_2$, start by ... $\mathbb{C}_3$, now let's use ... $\mathbb{C}_4$}'' still can help identify $\mathbb{C}_3$ and $\mathbb{C}_4$ as the correct solution.
This is quite similar to a question answering task~\cite{squad, hotpotqa}, which aims to identify the correct answer of the given question from a textual context.

Inspired by the above observations, we propose a Sequence Labeling based Question Answering (SLQA) method to mine NL-PL pairs in a PL-agnostic manner.
We first tokenize the text and code blocks of a post with a PL-independent tokenizer, which eliminates the characteristics of the PL itself.
Then we feed the title and the converted content into a sequence tagger.
The tagger assigns a label to each code-block token to indicate whether it belongs to the answer.
Moreover, we observe that many solutions provided by SO posts are composed of multiple code blocks interspersed with some explanatory text, as shown in Figure \ref{fig:intro_example}.
Therefore, we propose to apply the BIO tagging scheme~\cite{IOB} to predict multiple code blocks as answers instead of the binary scheme as in~\cite{staqc}.

Finally, we present Lang2Code, a parallel NL-PL corpus automatically mined with the SLQA model. 
The Lang2Code contains 1.4M NL-PL pairs over 6 PLs, which is so far the largest NL-PL corpus with the broadest coverage of PLs to our best knowledge.
We also conduct a downstream evaluation on a NL-to-code generation task, and the boosted performance by Lang2Code proves the high quality of Lang2Code.
The statistics and evaluation demonstrate that Lang2Code is a large-scale and high-quality NL-PL corpus and can greatly help model developments for tasks aiming to associate natural language with programming language.

The main contribution of this paper are:
\begin{itemize}
    \item {We propose the SLQA method to mine NL-PL pairs in a PL-agnostic manner.}
    \item {We propose to apply the BIO tagging scheme instead of the previous binary scheme to mine code solutions composed of multiple blocks.}
    \item {We conduct experiments on the current single-PL single-block benchmarks and a manually-annotated cross-PL multi-block benchmark, and the results demonstrate the effectiveness and transferability of our method.}
    \item {We further present Lang2Code, a large-scale and high-quality parallel NL-PL corpus automatically mined with SLQA, containing about \textbf{1.4M} NL-PL pairs over \textbf{6 PLs}.}
\end{itemize}

\section{Related Work}

\subsection{NL-PL Pair Mining}

Existing methods for mining NL-PL pairs first proposed to utilize the function's code-documentation pairs because they are naturally aligned and can be automatically mined from sources like GitHub.
For example, \citeauthor{csn}~\shortcite{csn} proposed the CodeSearchNet corpus by collecting function-level documentations on the GitHub.
\citeauthor{pymt5}~\shortcite{pymt5} proposed to translate documentation strings to Python codes with transformers~\cite{vaswani2017attention}.
\citeauthor{codebert}~\shortcite{codebert} proposed the CodeBERT model which is pre-trained with function-docstring pairs.
However, language used in documentations is fundamentally different from the general natural language.
A function's documentation is often written by the same author of the code and therefore does not provide much variability in terms of vocabulary \cite{csn}.

To address this issue, subsequent work focused on mining NL-PL pairs from linguistically richer but unstructured code forums from sources like SO.
For example, \citeauthor{codenn}~\shortcite{codenn} proposed a neural network to generate code summaries with posts that has only one code block.
\citeauthor{conala}~\shortcite{conala} proposed to extract structured features to explore Python and Java forums and present the automatically mined CoNaLa corpus.
\citeauthor{staqc}~\shortcite{staqc} proposed a bi-view hierarchical neural network and present the StaQC dataset.
However, existing methods usually required heavy PL-specific feature engineering or heavy PL-specific data annotations.
This poses great challenges for directly migrating them across multiple PLs, especially for niche PLs which lack annotated data.
To overcome these drawbacks, we propose a PL-agnostic way to predict multiple code blocks as a solution.
We treat the code contents as natural language and tokenize all the text and code blocks with a PL-independent tokenizer, which eliminates the specific characteristics of the PLs.
Then our model identifies the correct solution(s) of the question by understanding the global PL-agnostic textual context of the post.

\subsection{NL Context Understanding}

Recent progress of NLP techniques provided effective QA methods for comprehending query-related NL context.
For example, \citeauthor{bidaf}~\shortcite{bidaf} proposed a BiDAF model to capture interactions between contexts and queries.
\citeauthor{devlin2019bert}~\shortcite{devlin2019bert} and \citeauthor{liu2019roberta}~\shortcite{liu2019roberta} improved QA performance using  pre-trained language models.
However, these conventional QA models usually predicted only one contiguous answer span and are not suitable for answers consisting of consecutive code blocks interspersed with text blocks.
In comparison, we propose to address this issue with the sequence labeling mechanism~\cite{ner}.
We assign a BIO tag~\cite{IOB} to each code block to indicate whether it belongs to the solution of the question title. We elaborate on this method in Section \ref{subsec:manual_annotation}.

\begin{figure*}[t]
\centering
\includegraphics[width=.90\textwidth]{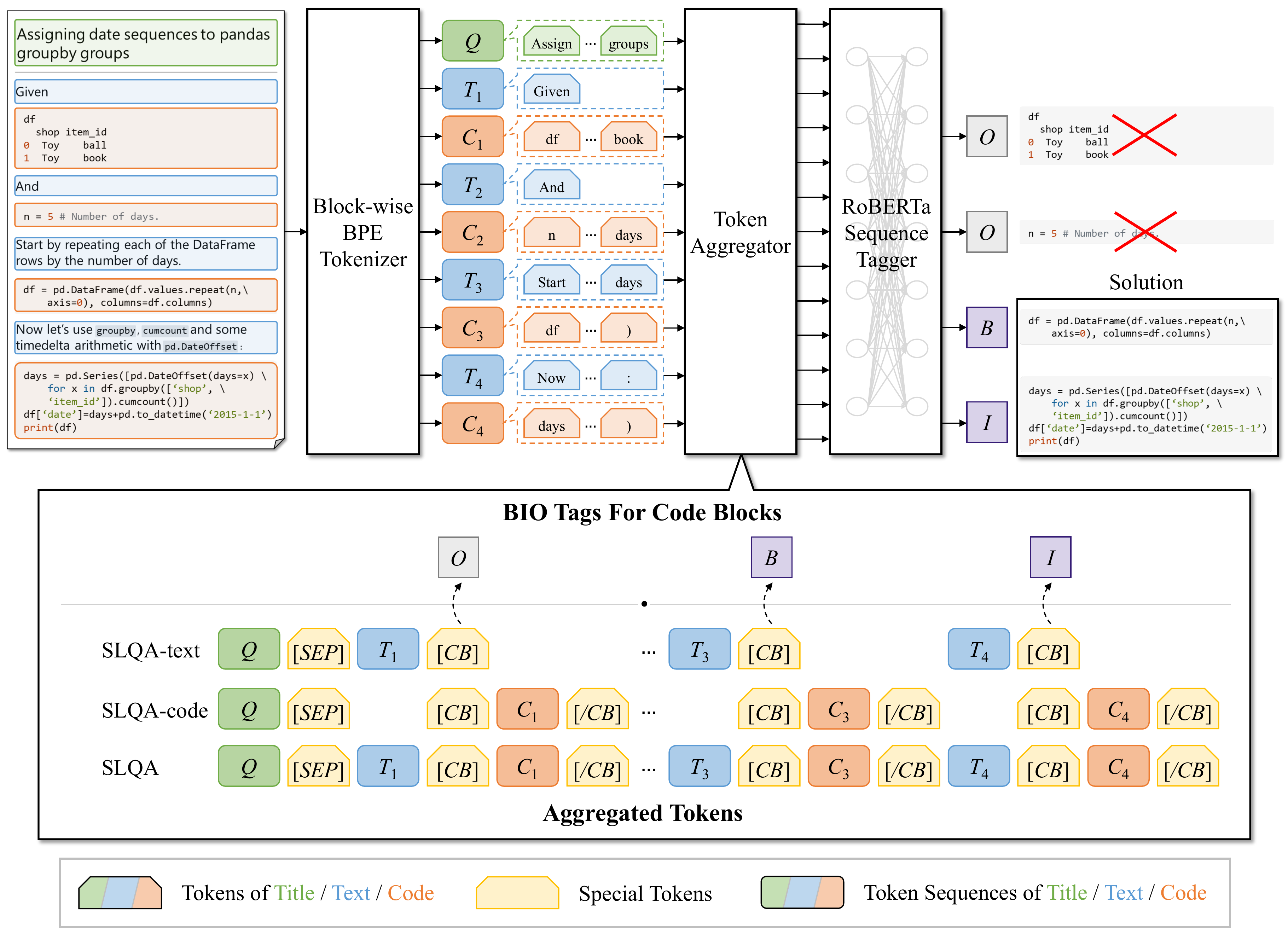}
\caption{The overall framework of SLQA and two variants: SLQA-text and SLQA-code. The SLQA-code and SLQA models use code block contents as input. However, the codes are tokenized with the same Byte-Pair Encoding (BPE)~\cite{bpe} tokenizer as the NL context. This means that these two models are still PL-agnostic.}
\label{fig:framework}
\end{figure*}

\section{Preliminaries}

In this section, we provide a task definition and describe how we annotated a cross-PL multi-block dataset for model development.

\subsection{Task Overview}

Given a Stack Overflow question title and its answer post accepted by the questioner, we aim at identifying all potential consecutive code blocks in the accepted answer that can solve the question. 
The solutions may contain one single code block or multiple code blocks interspersed with texts.
Following \citeauthor{staqc}~\shortcite{staqc}, we focus on "how-to-do-it" questions in SO, because their answer posts are more likely to contain complete code solutions.



\subsection{Manual Annotation}
\label{subsec:manual_annotation}

We first collected questions of six PLs from Stack Overflow's raw data dump\footnote{https://archive.org/details/stackexchange}: Python, Java, SQL, R, Git and Bash.
Then we manually selected how-to questions by three annotators.
We kept the questions with accepted answers, and for each question we only kept its title and the accepted answer post following~\cite{staqc}.
Next, we annotated the standalone code blocks\footnote{We ignored in-line code blocks as our pilot annotation study found that standalone blocks cover most of the solutions.} in the accepted answer with BIO tags.
Specifically, we use the following annotation protocol:
(1) non-solution blocks are labeled as "\textit{Outside} (\textit{O})";
(2) single-block solutions are labeled as "\textit{Begin} (\textit{B})"; 
(3) for multi-block solutions, the first code block is labeled as "\textit{B}" and all subsequent ones are labeled as "\textit{Inside} (\textit{I})".
This protocol was determined by three annotators through annotating a unified set of sample data and running five iterations of pilot study.
The average Cohen's kappa of the final iteration is around 0.785.
Finally, all data were exclusively divided to the three annotators for annotation.

\section{Our Approach}

The overall framework of our Sequence Labeling based Question Answering (SLQA) approach is shown in Figure \ref{fig:framework}.
First, we introduce the intuition in Section \ref{subsec:intuition}.
Then we introduce the tokenizer and the token aggregator in Section \ref{subsec:token_aggregator}.
Finally we introduce the sequence tagger in Section \ref{subsec:sequence_tagger}.

\subsection{Intuition}\label{subsec:intuition}

Consider an answer post with question title $\mathbb{Q}$ and a sequence of text and code blocks: $\{{\mathbb{T}_1}, {\mathbb{C}_1}, {\mathbb{T}_2}, {\mathbb{C}_2}... {\mathbb{T}_N}, {\mathbb{C}_N}\}$, where ${\mathbb{T}_{1:N}}$ and ${\mathbb{C}_{1:N}}$ denote the text and code blocks, respectively.
On one hand, we observe that the global textual context of ${\mathbb{T}_{1:N}}$ can provide \textit{PL-independent} information to identify the solution.
For example, in Figure \ref{fig:intro_example}, the words "\textit{Given}" and "\textit{And}" of the first two text blocks $\{\mathbb{T}_1, \mathbb{T}_2\}$ strongly indicate $\{\mathbb{C}_1,\mathbb{C}_2\}$ as prerequisites.
Then the keywords "\textit{Start by}" and "\textit{Now}" of $\{\mathbb{T}_3, \mathbb{T}_4\}$ suggest that ${\mathbb{C}_3}$ and ${\mathbb{C}_4}$ are the actual solution of the post.

On the other hand, the contents of the code blocks are intuitively informative.
However, previous work utilized \textit{PL-specific} parsers to pre-process the code~\cite{staqc}, which is not suitable for cross-PL transfer learning.
To overcome this drawback, our full model \textbf{SLQA} incorporates code contents by treating the code contents as natural language and using a unified \textit{PL-agnostic} tokenizer (Section \ref{subsec:token_aggregator}) to process all text and code blocks.

Surprisingly, we also discover that even without looking at the contents of all the code blocks, the remaining textual context ``\textit{Given $\mathbb{C}_1$ and $\mathbb{C}_2$, start by ... $\mathbb{C}_3$, now let's use ... $\mathbb{C}_4$}'' still can help identify the solution.
Therefore, we propose a pure-text variant of SLQA, denoted as \textbf{SLQA-text}.
Then another pure-code variant \textbf{SLQA-code} is naturally proposed, which is useful to investigate the effects of text and code separately, as shown in Figure \ref{fig:framework}.

\subsection{Tokenizer and Token Aggregator}\label{subsec:token_aggregator}

Given an answer post, we first use a unified tokenizer to tokenize the question title, the text blocks, and the code blocks.
However, the variance of different PL's keyword sets and the flexibility of the variables' names may lead to a large but sparse vocabulary.
We apply a Byte-Pair Encoding (BPE)~\cite{bpe} tokenizer to all blocks to alleviate data sparsity following~\cite{liu2019roberta}.
We denote the output token sequences of $\mathbb{T}_i$, $\mathbb{C}_i$ and $\mathbb{Q}$ as the italic symbols $T_i$, $C_i$ and $Q$, respectively.

Then we aggregate the above tokens to construct the sequence tagger's input (Figure \ref{fig:framework}).
First, we put $T_{1:N}$ and $C_{1:N}$ after $Q$ by the order of appearance.
Then, we insert two kinds of special tokens to the whole sequence:
(1) insert \texttt{[SEP]} after $Q$ to separate questions and contents~\cite{devlin2019bert};
(2) insert\texttt{[CB]} and \texttt{[/CB]} around each $C_i$ to distinguish codes from texts.
For SLQA-code, we remove all text tokens $T_{*}$.
For SLQA-text, we remove all code tokens $C_{*}$ and \texttt{[/CB]}s, and left \texttt{[CB]}s as placeholders of code blocks.

\subsection{Sequence Taggers}\label{subsec:sequence_tagger}

Inspired by QA methods~\cite{devlin2019bert,liu2019roberta}, we use a pre-trained language model to encode the input token sequence, such as RoBERTa~\cite{liu2019roberta} in our model.
We use its last hidden states $\mathbf{h}^{S}_{1:M}$ as the token representations of the input sequence.
However, conventional QA models only predict the begin and end positions of a \textit{contiguous} token span as the correct answer, thus failed to predict multiple code blocks interspersed with texts.
For example, in Figure \ref{fig:framework}, the answer tokens $C_3$ and $C_4$ are \textit{non-contiguous} because the token span is cut off by $T_4$ in the middle.
To address this issue, we propose to apply the sequence labeling method for answer prediction.
We predict a BIO tag for each \texttt{[CB]} token to find multiple consecutive code blocks as the solution, as shown in Figure \ref{fig:framework}.
Formally, we use a softmax classifier to predict the tag for the $i$-th block ($i\in\{1, 2, \dots, N\}$):
\begin{equation}
\begin{split}
    p(\mathbf{y}_i) &= \textrm{softmax}(\mathbf{W}\mathbf{h}_{m_i}^S +\mathbf{b})\\
    \textrm{tag}_{i} &= \arg\max_{k\in\{B,I,O\}}p(y_i=k)
\end{split}
\end{equation}
where $(\mathbf{W}, \mathbf{b})$ are the parameters of the classi, $m_i$ denotes the position of the $i$-th \texttt{[CB]} token.
We train our model with a cross-entropy loss function:
\begin{equation}
    \mathcal{L} = -\frac{1}{N}\sum_{i=1}^N \log p(y_i = \hat{y}_i)
\end{equation}
where $\hat{y}_{m_i}$ denotes the gold tag of the $i$-th block.

Note that \citeauthor{staqc}~\shortcite{staqc} also proposed to capture textual context for this task.
Our model essentially differs from theirs in that we apply a BIO tagging scheme to find multi-block solutions, whereas their binary scheme could only find single-block solutions.
Moreover, we model the post content as a whole sequence and capture its global textual context, while their methods only incorporated the local textual context around the code blocks.
We show that the global textual context provides more information than the local context in Section \ref{sec:exp_staqc}.

\section{Experiments}
\label{sec:experiments}

\subsection{Datasets}
We evaluate our methods on two datasets, which are referred to as \textit{StaQC-human} and \textit{Lang2Code-human} in the following sections.

\subsubsection{StaQC-human}

StaQC~\cite{staqc}  consists of 4,883 Python and 3,637 SQL manually annotated question-code (QC) pairs.
Each code block is annotated with 0 or 1 to indicate whether it solves the corresponding question.
This means that StaQC discards all multi-block joint solutions.
The original training, validation and test sets are split at the code block level and therefore not applicable to our post-level approach.
Moreover, two consecutive code blocks from the same post may be placed into different subsets.
Since they share the same text block between them, this placement may cause potential data leakage.
Therefore, we reassemble the code blocks into posts and randomly split the fully-annotated posts into new training, validation and test sets.
Then we transform the $\{1, 0\}$ labels into $\{B, O\}$ tags.
We denote the new single-block dataset as StaQC-human.
Table \ref{tab:staqc_statistics} shows the statistics of the single-block StaQC-human dataset.




\begin{table}[!t]
\centering
\caption{Statistics of the StaQC-human dataset. Non. stands for non-solution blocks.}
\begin{tabular}{llccc}
\toprule
\multirow{2.4}{*}{Subset} & \multirow{2.4}{*}{Lang.}  & \multirow{2.4}{*}{\#Post} & \multicolumn{2}{c}{\#Block}\\
\cmidrule(l){4-5}
                      &        &       & Single  & Non. \\
\midrule
Train & Python & 1,139 & 1,354 & 1,592 \\
Valid & Python & 142 & 181 & 197 \\
Test & Python & 143 & 164 & 204 \\
\midrule
Train & SQL    & 1,056 & 1,632 & 1,016 \\
Valid & SQL & 132 & 201 & 124 \\
Test & SQL & 133 & 190 & 135 \\
\bottomrule
\end{tabular}
\label{tab:staqc_statistics}
\end{table}


\subsubsection{Lang2Code-human}

\begin{table}[t]
\caption{Statistics of the Lang2Code-human dataset. \textit{Multi.} stands for code blocks that form multi-block solutions, and \textit{Non.} stands for non-solution blocks.}
\centering
\begin{tabular}{llcccc}
\toprule
\multirow{2.4}{*}{Subset} & \multirow{2.4}{*}{Lang.}  & \multirow{2.4}{*}{\#Post} & \multicolumn{3}{c}{\#Block}\\
\cmidrule(l){4-6}
                      &        &       & Single  & Multi.& Non. \\
\midrule
Train                 & Python & 1,058 & 1,328   & 423   & 1,131 \\
\midrule
Valid                 & Python & 132   & 185     & 58    & 140   \\
\midrule
\multirow{6}{*}{Test} & Python & 133   & 178     & 44    & 149   \\\
                      & Java   & 53    & 47      & 68    & 23    \\
                      & SQL    & 70    & 91      & 33    & 60    \\
                      & R      & 80    & 88      & 44    & 84    \\
                      & Git    & 75    & 63      & 107   & 41    \\
                      & Bash   & 69    & 87      & 63    & 55    \\
\bottomrule
\end{tabular}
\label{tab:lang2code_dataset}
\end{table}

%
%
%
We construct the Lang2Code-human dataset with our manually-annotated data as mentioned in Section \ref{subsec:manual_annotation}.
We focused on Python as the primary PL.
The Python posts are used as the training, validation sets and part of the test set.
They contain both randomly-sampled and hits-weighted sampled data to jointly represent the distributions of all SO posts and the popular ones.
Posts of other five PLs (Java, SQL, R, Git, Bash) are randomly sampled and used as the test set for cross-PL performance benchmarking.
The statistics of the multi-block Lang2Code-human dataset is shown in Table \ref{tab:lang2code_dataset}.

\begin{table*}[t]
\centering
\caption{Performance of SLQA and previous method on StaQC-human's Python and SQL test sets. The
best scores are in bold and the second-best scores are underlined. $\dagger$ denotes cross-PL performance.}
\begin{tabular}{lcccccccccc}
\toprule
\multirow{2.4}{*}{Method} & \multicolumn{5}{c}{Python     }     & \multicolumn{5}{c}{SQL     }        \\
\cmidrule(lr){2-6} \cmidrule(lr){7-11}
                        & Train  & Prec. & Rec. & $F_1$.  & Acc. & Train  & Prec. & Rec. & $F_1$.  & Acc. \\
\midrule
Select-First            & Python & 64.3  & 56.1 & 59.9 & 66.6 & SQL    & 73.7  & 51.3 & 60.5 & 60.6 \\
Select-All              & Python & 44.6  & \textbf{100}  & 61.7 & 44.6 & SQL    & 58.8  & \textbf{100}  & 74.1 & 58.8 \\
Text-HNN                & Python & 71.7  & 79.5 & 75.4 & 76.8 & SQL    & 76.5  & 86.6 & 81.2 & 76.5 \\
Code-HNN                & Python & 71.7  & 77.8 & 74.6 & 76.4 & SQL    & 75.3  & 88.8 & 81.5 & 76.3 \\
BiV-HNN                 & Python & 76.2  & 83.8 & 79.8 & 81.0 & SQL    & 83.2  & 96.0 & 89.1 & 86.1 \\
BiV-RoBERTa             & Python & 75.8  & 76.1 & 75.9 & 77.7 & SQL    & 82.9  & 91.1 & 86.8 & 83.4 \\
SLQA-text               & Python & 78.5  & 86.0 & \underline{82.1} & \underline{83.0} & SQL    & 86.3  & 94.0 & \underline{90.0} & \underline{87.7} \\
SLQA-code               & Python & 72.8  & 87.0 & 79.3 & 79.1 & SQL    & 84.2  & 89.0 & 86.5 & 83.5 \\
SLQA                    & Python & \textbf{85.3}  & \underline{90.2} & \textbf{87.7} & \textbf{88.4} & SQL    & 85.2  & \underline{97.7} & \textbf{91.0} & \textbf{88.4} \\
\midrule
SLQA-text$^\dagger$     & SQL    & 74.6  & 90.1 & 81.6 & 81.9 & Python & \underline{88.0}  & 82.8 & 85.3 & 83.1 \\
SLQA-code$^\dagger$     & SQL    & \underline{84.5}  & 24.3 & 37.7 & 63.6 & Python & 77.3  & 80.1 & 78.7 & 74.0 \\
SLQA$^\dagger$          & SQL    & 79.5  & 82.3 & 80.9 & 82.3 & Python & \textbf{90.4}  & 88.3 & 89.3 & 87.2 \\
\bottomrule
\end{tabular}
\label{tab:staqc-Python}
\end{table*}

\subsection{Experimental Settings}


We tune the hyper-parameters on the validation sets with earlystopping patience of 5. 
We load the pre-trained RoBERTa-Large parameters\footnote{https://huggingface.co/roberta-large} and fine-tune them during training.
We use the Adam~\cite{kingma2014adam} optimizer to train our model with the learning rate of $10^{-5}$ and 500 warmup steps.
We set the weight decay to $0.01$.
We report the precision, recall and $F_1$ scores of the predict solutions on both datasets.
We also report the accuracy on StaQC-human because of its binary labels.
We train our model for five times with different random seeds and report the average scores.




\subsection{Baselines}

\begin{figure}[t]
\centering
\includegraphics[width=.35\textwidth]{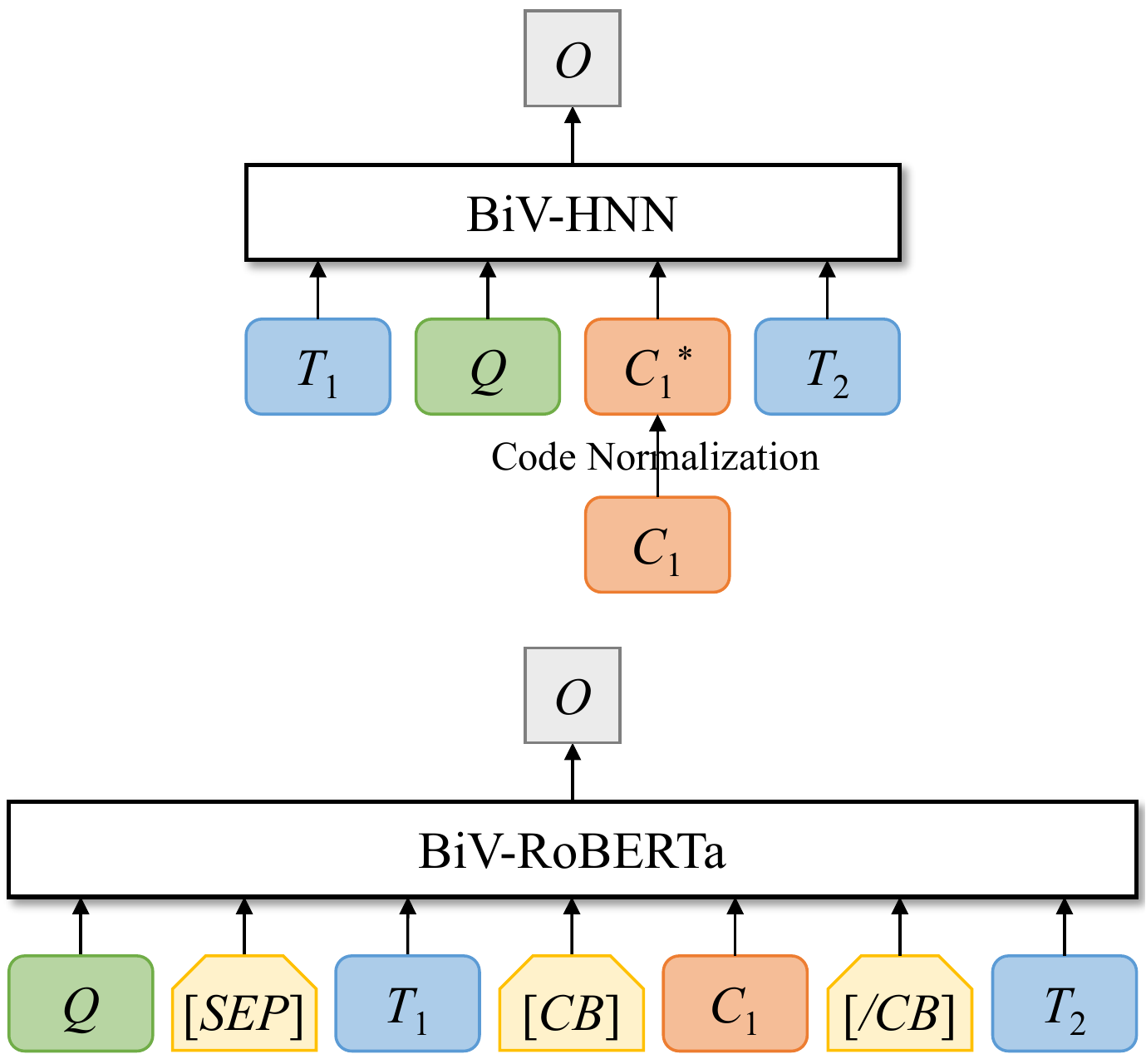}
\caption{The overall framework of BiV-HNN and BiV-RoBERTa. BiV-HNN is a GRU based model that takes normalized code as input, where as BiV-RoBERTa is a transformer based model takes raw code as input.}
\label{fig:BiV-RoBERTa}
\end{figure}

We compare our model with the following baselines:
(1) \textbf{Select-All} and \textbf{Select-First} are two commonly used heuristics~\cite{staqc}.
Select-All treats all code blocks in a post as standalone solutions, i.e., all code blocks are labeled as "\textit{B}". Select-First only chooses the first code block as the standalone solution, which is equivalent to labeling the first code block as "\textit{B}" and the rest as "\textit{O}".
(2) \textbf{BiV-HNN}~\cite{staqc} learns the semantic representation of a code block from token level to block level and incorporates the local textual context around each code block.
It has two main variants, \textbf{Text-HNN} and \textbf{Code-HNN}, which discard code contents and textual context, respectively.
Although StaQC contains data of two PLs, these models are not available for cross-PL experiments because they used PL-specific vocabularies.
Since StaQC-human is re-split at the post level, we re-train BiV-HNN and its variants on StaQC-human with the official implementation and report the post-level scores.
(3) \textbf{BiV-RoBERTa} is similar to BiV-HNN, but uses the RoBERTa instead of the HNN to capture the code contents and local textual context (Figure \ref{fig:BiV-RoBERTa}).

\subsection{StaQC-human Results}
\label{sec:exp_staqc}

We conduct single-PL and cross-PL experiments on StaQC-human Python dataset and report the results in Table \ref{tab:staqc-Python}.
From the results we have the following observations:


\textbf{Single-PL}:
(1) SLQA significantly outperforms previous Select-First, Select-All and BiV-HNN methods. It indicates that our model can effectively capture global textual context and code contents and improve the performance.
(2) SLQA outperforms the BiV-RoBERTa model.
The two models use the same RoBERTa encoder, the only difference is that SLQA uses global contextual information while the other uses local contextual information.
It indicates the effectiveness of the global textual context for NL-PL understanding.
Additionally, BiV-RoBERTa performs worse than BiV-HNN.
It shows that the PL-specific code normalization (Figure \ref{fig:BiV-RoBERTa}) provides major contribution to the performance.
SLQA performs even better than BiV-HNN without code normalization, which indicates that the improvements come primarily from the global textual context rather than the pre-trained RoBERTa.
(3) SLQA-text also outperforms the BiV-HNN model. 
Since SLQA-text has no access to the code contents but still brings improvements, this result further demonstrates the effectiveness of our method to capture global textual information.

\begin{table}[t]
\centering
\caption{
Performance of SLQA and heuristics on Lang2Code-human test set, including all the six PLs. The best scores are in bold and the second-best scores are underlined.}
\begin{tabular}{lcccccc}
\toprule
\multirow{2.4}{*}{Method} & \multicolumn{6}{c}{$F_1$} \\ \cmidrule(l){2-7} 
\multicolumn{1}{c}{} & Python & Java & SQL & R & Git & Bash \\ 
\midrule
Select-First & 49.8 & 37.5 & 40.2 & 45.2 & 30.2 & 41.3 \\
Select-All & 63.4 & 44.1 & 63.7 & 54.5 & 41.1 & 55.9 \\ 
\midrule
SLQA-text & 69.7 & \underline{53.8} & \underline{70.0} & 64.0 & \underline{55.4} & \underline{65.8} \\
SLQA-code & \underline{70.6} & 50.0 & 59.8 & \underline{64.7} & 48.0 & 61.5 \\
SLQA & \textbf{75.7} & \textbf{62.9} & \textbf{76.1} & \textbf{66.5} & \textbf{58.7} & \textbf{70.1} \\ 
\bottomrule
\end{tabular}
\label{tab:lang2code-SLQA}
\end{table}

\begin{table}[t]
\centering
\caption{
The statistics of the Lang2Code corpus. 
\#Solution Multi. refers to the number of NL-PL pairs that PL consists of multiple code blocks.
\#Blocks Multi. refers to the total number of code blocks combined in those NL-PL pairs that consist of multiple code blocks.}
\begin{tabular}{lrrrrr}
\toprule
\multirow{2}{*}{Lang.} & \multicolumn{1}{c}{\multirow{2}{*}{\#Pairs}} &  \multicolumn{2}{c}{\#Solutions} & \multicolumn{2}{c}{\#Blocks} \\ \cmidrule(l){3-6} 
 &  & \multicolumn{1}{c}{Single} & \multicolumn{1}{c}{Multi.} & \multicolumn{1}{c}{Single} & \multicolumn{1}{c}{Multi.} \\ 
\midrule
Python & 572,587 & 537,768 & 34,819 & 537,768 & 81,574 \\
Java & 322,731 & 295,920 & 26,811 & 295,920 & 62,487 \\
SQL & 234,939 & 226,548 & 8,391 & 226,548 & 19,040 \\
R & 195,685 & 181,862 & 13,823 & 181,862 & 33,229 \\
Git & 27,470 & 24,164 & 3,306 & 24,164 & 8,549 \\
Bash & 45,099 & 43,104 & 1,995 & 43,104 & 4,801 \\
\midrule
Total & 1,398,511 & 1,309,366 & 89,145 & 1,309,366 & 209,680 \\
\bottomrule
\\
\end{tabular}
\label{tab:statstics_lang2code}
\end{table}

\textbf{Cross-PL}:
(1) SLQA-text achieves comparable results to the single-PL BiV-HNN model.
It indicates that the global textual context provides PL-agnostic supplementary information and allows for transfer learning across different PLs.
(2) SLQA-code performs well on single-PL task but poorly on cross-PL task, even worse than the heuristic baselines on the Python test set. 
It validates the importance of textual information to cross-PL transfer learning.
(3) Cross-PL SLQA outperforms single-PL BiV-HNN and achieves comparable results to single-PL SLQA-text.
It indicates that our model can effectively incorporate the code contents and textual context in a PL-agnostic manner and improve the cross-PL transfer learning performance.

\subsection{Lang2Code-human Results}
\label{sec:exp_lang2code}

We present the experimental results on the cross-PL multi-block Lang2Code dataset in Table \ref{tab:lang2code-SLQA}.
We can observe that:
(1) The SLQA scores on the Python test set are significantly lower than StaQC-human's single-PL scores.
It demonstrates that predicting multi-block solutions is a more challenging task than predicting single-block solutions.
(2) SLQA-code performs poorly on cross-PL learning except for Python-to-R learning.
It indicates that only focusing on code contents is not effective enough for cross-PL learning because the language styles varies greatly between different PLs, unless the two PLs have similar styles like Python and R.
(3) SLQA-text significantly outperforms heuristic methods.
It indicates that the global textual context can provide critical PL-agnostic information to identify the correct solution from an SO post.
Our model effectively captures the global textual context and improve the performance of cross-PL transfer learning.
(4) The full SLQA model significantly outperforms both the variants and the heuristics.
It indicates that SLQA effectively captures global textual context and code contents of a post and allows for reliable transfer learning across multiple PLs.


\section{Lang2Code: An automatically mined NL-PL pairs dataset}\label{sec:lang2code}

In this section, we present Lang2Code, a large-scale and high-quality parallel NL-PL corpus automatically mined using our SLQA model. 
It contains about \textbf{1.4M} NL-PL pairs over \textbf{6 PLs}, as shown in Table \ref{tab:statstics_lang2code}.
We will introduce the construction pipeline in Section \ref{subsec:construct}, the statistics in Section \ref{subsec:statistics}, and the downstream evaluation of data quality in Section \ref{subsec:eval}, respectively.

\subsection{Construction Pipeline}\label{subsec:construct}

\begin{figure}[t]
\centering
\includegraphics[width=.43\textwidth]{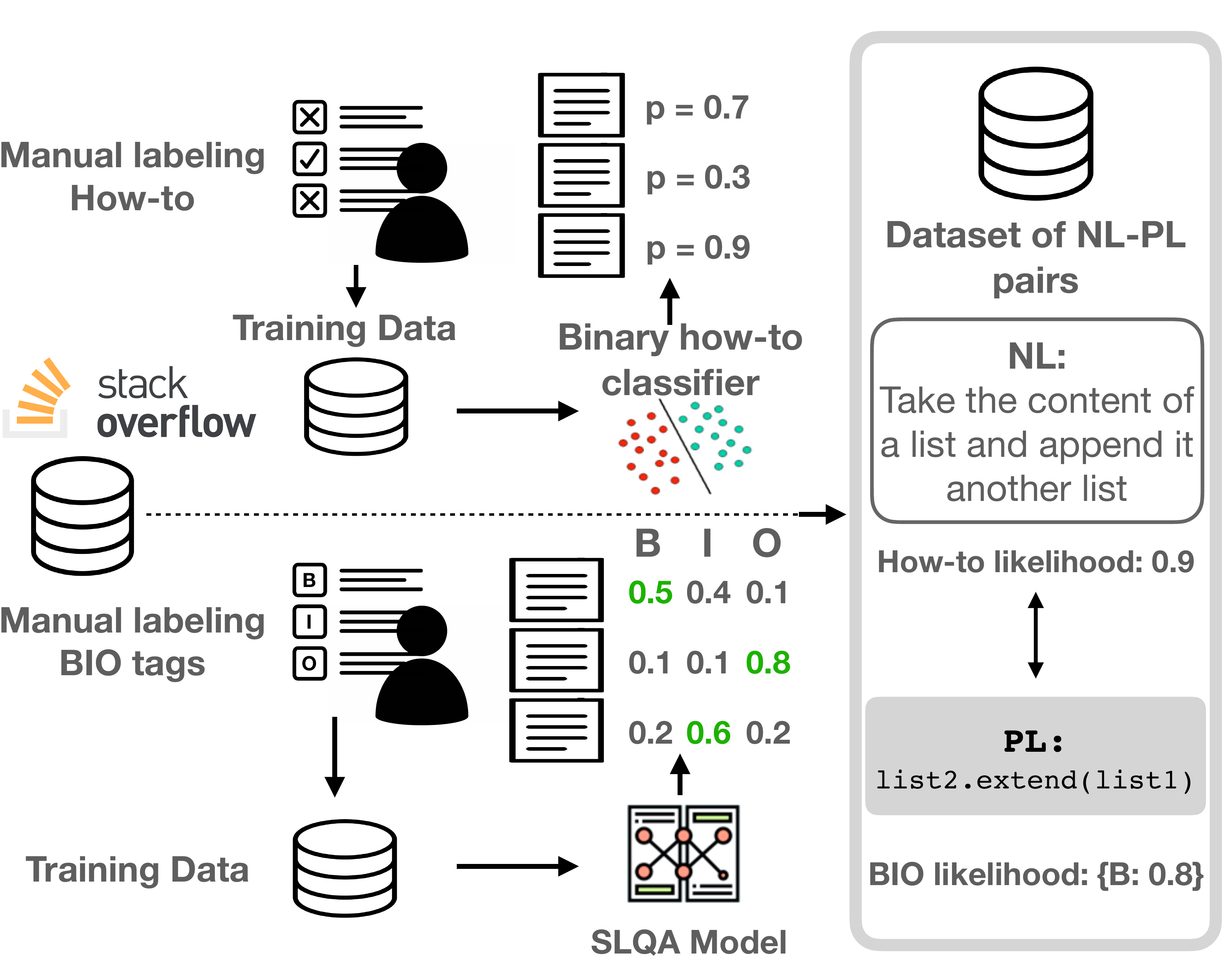}
\caption{The overall pipeline of automatically constructing the Lang2Code dataset.}
\label{fig:pipeline}
\end{figure}

\begin{figure}[t]
\centering
\includegraphics[width=.42\textwidth]{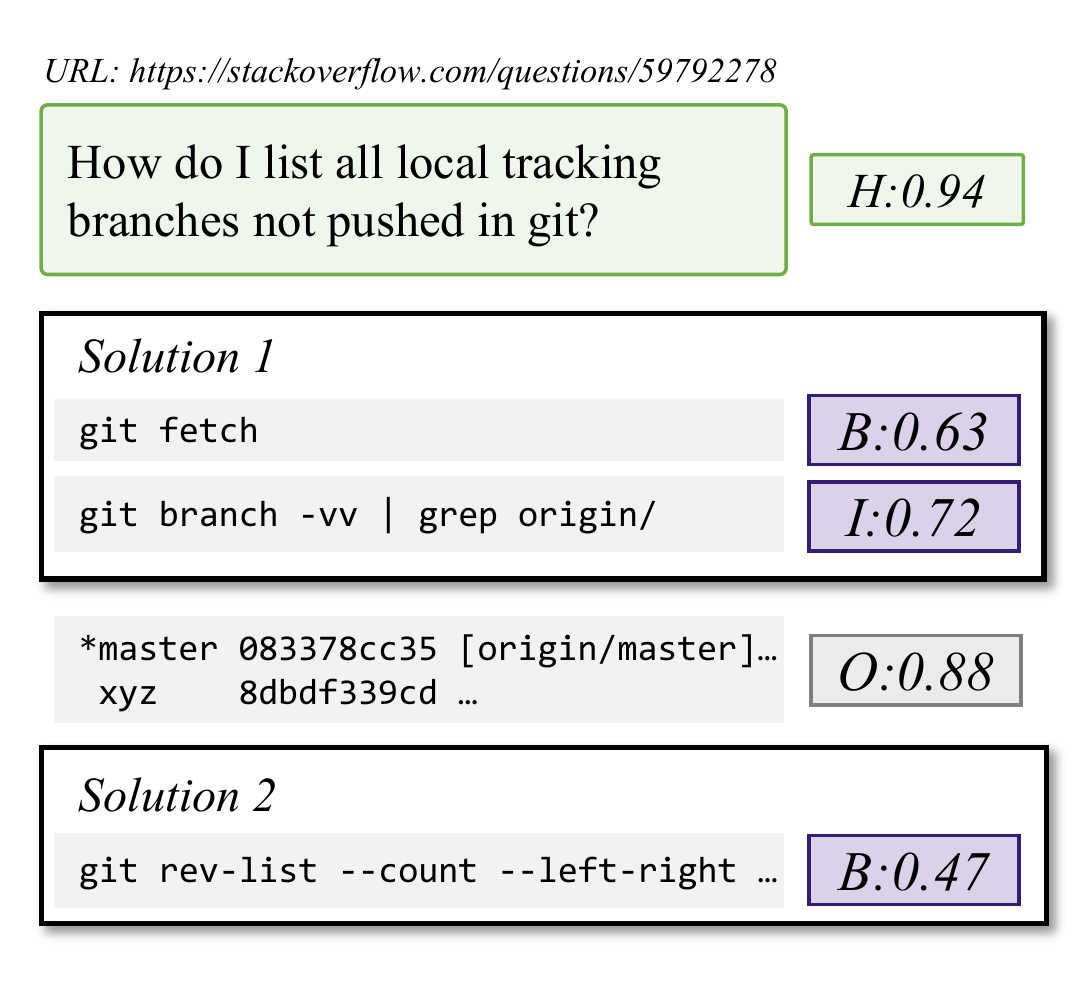}
\caption{An illustration of the Lang2Code Corpus. In this SO post, two solutions corresponding to the question title were obtained, resulting in two NL-PL pairs. We include both the how-to likelihood and the BIO likelihoods in Lang2Code.}
\label{fig:case_study}
\end{figure}

\begin{table*}[t]
\centering
\caption{Overview of NL-PL datasets auto-mined from Stack Overflow. The largest size are in bold.}
\begin{tabular}{lllllllll}
\toprule
\multicolumn{1}{l}{Datasets} & \multicolumn{1}{l}{Python} & \multicolumn{1}{l}{Java} & \multicolumn{1}{l}{SQL} & \multicolumn{1}{l}{R} & \multicolumn{1}{l}{Git} & \multicolumn{1}{l}{Bash} & \multicolumn{1}{l}{C\#} & \multicolumn{1}{l}{Total} \\ 
\midrule
\multicolumn{1}{l}{CodeNN \cite{codenn}} & \multicolumn{1}{c}{/} & \multicolumn{1}{c}{/} & \multicolumn{1}{c}{33k} & \multicolumn{1}{c}{/} & \multicolumn{1}{c}{/} & \multicolumn{1}{c}{/} & \multicolumn{1}{c}{53k} & \multicolumn{1}{l}{86k} \\ 
\multicolumn{1}{l}{StaQC \cite{staqc}} & \multicolumn{1}{c}{148k} & \multicolumn{1}{c}{120k} & \multicolumn{1}{c}{/} & \multicolumn{1}{c}{/} & \multicolumn{1}{c}{/} & \multicolumn{1}{c}{/} & \multicolumn{1}{c}{/} & \multicolumn{1}{l}{268k} \\ 
\multicolumn{1}{l}{SO-DS \cite{sods}} & \multicolumn{1}{c}{12k} & \multicolumn{1}{c}{/} & \multicolumn{1}{c}{/} & \multicolumn{1}{c}{/} & \multicolumn{1}{c}{/} & \multicolumn{1}{c}{/} & \multicolumn{1}{c}{/} & \multicolumn{1}{l}{12k} \\ 

\multicolumn{1}{l}{CoNaLa \cite{conala}} & \multicolumn{1}{c}{\textbf{601k}} & \multicolumn{1}{c}{/} & \multicolumn{1}{c}{/} & \multicolumn{1}{c}{/} & \multicolumn{1}{c}{/} & \multicolumn{1}{c}{/} & \multicolumn{1}{c}{/} & \multicolumn{1}{l}{601k} \\ 
\midrule
\multicolumn{1}{l}{Lang2Code} & \multicolumn{1}{c}{573k} & \multicolumn{1}{c}{\textbf{323k}} & \multicolumn{1}{c}{\textbf{235k}} & \multicolumn{1}{c}{\textbf{196k}} & \multicolumn{1}{c}{\textbf{27k}} & \multicolumn{1}{c}{\textbf{45k}} & \multicolumn{1}{c}{/} & \multicolumn{1}{l}{\textbf{1399k}} \\ 
\bottomrule
 \\
\end{tabular}
\label{tab:Dataset-Comparison}
\end{table*}

The overall pipeline is shown in Figure \ref{fig:pipeline}.
In Section \ref{subsec:manual_annotation}, we obtained 2,351 manually annotated tags from Lang2Code for the six languages, concomitant with these tags are also binary labels indicating whether the question is a How-to question. 
Using this data, we fine-tuned BERT \footnote{https://huggingface.co/bert-base-cased} to train a binary classifier, 
obtaining 85.2\% precision, 95.1\% recall and 89.9\% $F_1$ score.
Next we used this classifier with a threshold of 0.5 to filter Stack Overflow posts pertaining to the six PLs we annotated in Section \ref{subsec:manual_annotation}. 
We filtered 1,000,888 single-block answer posts and 799,104 multi-block answer posts and kept 540,352 and 676,128 how-to questions respectively.
After that, for single-block answer posts, we follow previous work \cite{codenn, conala, staqc} to treat the only code block as the solution to the question title. 
Thus we can have a NL-PL pair out of each single-code answer posts.
And for posts contain multiple code blocks, we applied SLQA trained on Lang2Code-human to the question answer pair and assigned BIO tags to every code blocks.
As discussed in Section \ref{subsec:manual_annotation}, using BIO tags we can segregate multiple code-block solutions as the ones with "\textit{B}" followed by one or more  "\textit{I}" tags, whereas single code-block solutions are ones with just a "\textit{B}" tag followed by "\textit{O}" tags or is the last code block. We pair each code-block solution with the question title to construct an NL-PL pair. For example, in Figure \ref{fig:case_study}, We use Solution 1 and Solution 2 to pair with the question title separately to obtain two NL-PL pairs.

\subsection{Statistics of Lang2Code}\label{subsec:statistics}

We have mined 858,159 NL-PL pairs from multi-code answer posts, together with the NL-PL pairs obtained from single-code answer posts, making a total of 1,398,511 NL-PL pairs in total. 
Table \ref{tab:statstics_lang2code} presents the detailed statistics of the Lang2Code corpus.
To the best of our knowledge, this is the largest NL-PL corpus mined from Stack Overflow so far, and the first corpus containing code solutions that are composed of multiple code blocks.
We will in the future publish the Lang2Code corpus and open source all related code for future research.
Moreover, we compare the amount of data for each subset of programming languages in the existing NL-PL datasets with Lang2Code.
Table \ref{tab:Dataset-Comparison} shows the comparison results.
From the results we can observe that Lang2Code is the dataset with the broadest coverage of programming languages, and on every PL language it collects more NL-PL pairs than previous datasets. 
In addition, Lang2Code first presents Stack Overflow-based dataset on Git, R, and Bash, which greatly helps develop data-hungry models for these niche PLs in future NL-PL research.

\subsection{Downstream Evaluation}\label{subsec:eval}

To validate the data quality of the auto-mined Lang2Code corpus, we conduct a downstream evaluation on a representative NL-to-code generation task~\cite{incorporating_conala,norouzi2021code}.
The task is defined as translating a natural language query (usually a description of a functionality) to  corresponding code snippet(s).
For example, given the intent to ``\textit{generate random integers between 0 and 9}'', one would expect the Python code snippet ``\texttt{import random; random.randint(0, 9)}''.
The performance on this task represents the ability to model the association between NL and PL.
For evaluation purposes, we choose the widely-used CoNaLa-human~\cite{conala} as the benchmark dataset.
It covers real-world English queries about Python with
diverse intents.
The dataset contains 2,179 training, 200 dev and 500 test human-annotated samples.
We use the same evaluation metric as the CoNaLa benchmark, corpus-level BLEU calculated on target code outputs in the test set.

\begin{table}[t]
\caption{Performance of CoNaLa, StaQC and Lang2Code pre-trained model on the code generation task. The best score is in bold. All the scores are produced without using the distribution sampling \cite{incorporating_conala} and re-ranking techniques \cite{rerank}.}
\centering
\begin{tabular}{ll}
\toprule
\multicolumn{1}{l}{Method} & \multicolumn{1}{c}{BLEU} \\ 
\midrule
\multicolumn{1}{l}{Man} & \multicolumn{1}{c}{27.20} \\ 
\multicolumn{1}{l}{Man+CoNaLa-mined} & \multicolumn{1}{c}{28.14} \\ 
\multicolumn{1}{l}{Man+StaQC-mined} & \multicolumn{1}{c}{28.29} \\ 
\midrule
\multicolumn{1}{l}{Man+Lang2Code-mined} & \multicolumn{1}{c}{29.83} \\
\multicolumn{1}{l}{Man+Lang2Code-mined-clean} & \multicolumn{1}{c}{\textbf{30.68}} \\

\bottomrule
\end{tabular}
\label{tab:Downstream-CoNaLa}
\end{table}

\begin{table*}[t]
\centering
\caption{Examples of code snippets generated by models pre-trained with CoNaLa, StaQC, and Lang2Code. The italic comments are not generated by the models but added manually by authors.}
\begin{tabular}{ll|l}
\toprule
\multicolumn{1}{l}{\textbf{Intent}} & 
\multicolumn{1}{l}{1. write line ``Hello'' to file ``somefile.txt''}  &
\multicolumn{1}{l}{
\begin{tabular}[c]{@{}l@{}}
2. pandas: delete rows in dataframe ``df'' based on \\
\ \ \ \ multiple columns values
\end{tabular} }\\ 
\midrule
\multicolumn{1}{l}{\textbf{Gold}} & 
\multicolumn{1}{l}{
\begin{tabular}[c]{@{}l@{}}
\texttt{with open('somefile.txt', 'a') as the\_file:}\\
\ \ \ \ \ \ \ \ \texttt{the\_file.write('Hello\textbackslash{}\textbackslash{}n')}
\end{tabular}} &
\multicolumn{1}{l}{
\begin{tabular}[c]{@{}l@{}}
\texttt{df.set\_index(list('BC'))}\\
\ \ \ \ \texttt{.drop(tuples, errors='ignore')}\\
\ \ \ \ \texttt{.reset\_index()}
\end{tabular}}
\\ 
\midrule
\multicolumn{1}{l}{\textbf{CoNaLa}} &
\multicolumn{1}{l}{\texttt{f.write('Hello')}} &
\multicolumn{1}{l}{\texttt{df.drop([df], axis=1)} \textit{\# error, delete df itself}}\\
\midrule
\multicolumn{1}{l}{\textbf{StaQC}} & \multicolumn{1}{l}{
\begin{tabular}[c]{@{}l@{}}
\texttt{with open ( 'somefile . txt ', 'a') as myfile:} \\
\ \ \ \ \ \ \ \ \texttt{myfile.write ( 'Hello')}
\end{tabular}} &
\multicolumn{1}{l}{\texttt{df.apply(lambda x:x.tolist(), axis=1)}}\\ 
\midrule
\multicolumn{1}{l}{\textbf{Lang2Code}} & \multicolumn{1}{l}{
\begin{tabular}[c]{@{}l@{}}
\texttt{with open ( 'somefile . txt ', 'a') as f:} \\
\ \ \ \ \ \ \ \ \texttt{f.write ( 'Hello')}
\end{tabular}}  &
\multicolumn{1}{l}{\texttt{df.drop('str\_2', axis=1, inplace=True)}}\\ 
\bottomrule
\end{tabular}
\label{tab:Code-Translation-Example}
\end{table*}

\subsubsection{Baselines}

\citeauthor{yin-neubig-2018-tranx}~\cite{yin-neubig-2018-tranx} proposed a transition based LSTM model named TranX for this task. 
Furthermore, \citeauthor{incorporating_conala}~\cite{incorporating_conala} proposed to incorporate mined NL-PL pairs as external knowledge.
Specifically, they first pre-trained TranX on the top100k auto-mined CoNaLa NL-PL pairs (sorted by confidence score).
Then the model is fine-tuned on CoNaLa-human training set, and finally evaluated on the CoNaLa-human test set.
To investigate the data quality of the auto-mined Lang2Code data, we conduct experiments of replacing the mined CoNala pairs with StaQC~\cite{staqc} and Lang2Code mined pairs.
We compare the performance of the following settings:
\begin{itemize}
    \item {\textbf{Man}: The model is solely trained with the CoNaLa-human training set without pre-training.}
    \item {\textbf{Man+CoNaLa-Mined}: The model is first pre-trained with \textit{top 100k mined CoNaLa pairs}, and then fine-tuned with the CoNaLa-human training set.}
    \item {\textbf{Man+StaQC-Mined}: We replace the CoNaLa pre-train data with 100k \textit{StaQC mined} pairs. Since StaQC has no confidence score, we randomly sample 100k pairs from the whole corpus.}
    \item {\textbf{Man+Lang2Code-Mined}: We replace the CoNaLa pre-train data with top 100k \textit{Lang2Code-mined} pairs compared with Man+CoNaLa-Mined.}
    \item {\textbf{Man+Lang2Code-Mined-clean}: 
    Following CoNaLa's pre-processing procedure, we clean the top 100k Lang2Code mined pairs by removing the \texttt{import} statements because they are considered as contexts rather than code snippets~\cite{conala}.
    We also remove code snippets longer than 100 characters to avoid distribution bias of code length. 
    We pre-train the model with the \textit{cleaned} Lang2Code pairs and then fine-tune it with the CoNaLa-human training set.}
\end{itemize}

\subsubsection{Results}

We report the BLEU scores of the five settings in Table \ref{tab:Downstream-CoNaLa}.
From the results we can observe that pre-training with Lang2Code mined pairs significantly boosts the performance. It indicates that our Lang2Code corpus contains less noise and the data quality is higher than the mined CoNaLa and StaQC pairs.
With higher-quality pre-training data, the model is able to understand the association between NL and PL more accurately.


\subsubsection{Case Study}
Table \ref{tab:Code-Translation-Example} shows the comparison of the models pre-trained with CoNaLa, StaQC and Lang2Code mined pairs on two example intents.
First, we can observe that the CoNaLa model tends to generate short or incorrect code snippets compared with the other two models.
We hypothesize that this is because the code snippets in CoNaLa are mined at the line level.
Specifically, the CoNaLa code snippets only contains 1.07 lines in average~\cite{ncs_revisit}.
In contrast, the StaQC and Lang2Code's block-level snippets contains 9.84 and 8.27 lines in average, respectively.
Therefore, for the CoNaLa pre-trained model, the difficulty to generate multiple lines of code is greatly increased.
Moreover, short code snippets usually have incomplete function, which may lead to generation results with obvious errors.
This result indicates that CoNaLa mined pairs are more likely to miss important code lines than StaQC and Lang2Code, which is harmful to the downstream performance.
Second, from the second intent we observe that the model pre-trained on StaQC generates semantically unrelated code snippets to the intent.
We hypothesize that the StaQC pairs are mined only using the local textual context around each code block.
In contrast, our Lang2Code pairs are mined with the global textual context of the whole answer post.
As mentioned in Section \ref{sec:exp_staqc}, the global textual context is more effective than the local context in understanding the association between NL and PL.
This result indicates that the StaQC mined pairs are more noisy than Lang2Code because the NL and PL within a pair are less associated.
The above results demonstrate the higher data quality of Lang2Code than StaQC and CoNaLa, which is more beneficial to future NL-PL research.

\section{Conclusion}
In this paper, we propose a Sequence Labeling based Question Answering (SLQA) approach to construct NL-PL pairs in a PL-agnostic way.
We propose a model that can produce multiple code blocks as solutions of a post's question, which is achieved by using BIO sequence tagging.
We also propose to incorporate the global textual context as PL-independent supplementary information.
To validate the capacity of our method, we manually annotate a challenging cross-PL multi-block dataset, named Lang2Code-human.
Substantial experiments on the single-PL single-block StaQC-human and our Lang2Code-human benchmarks demonstrate the effectiveness and cross-PL transferability of our method.
Finally, we present Lang2Code, the largest-to-date NL-PL corpus to the best of our knowledge, containing over 1.4 million pairs spanning six PLs.
Under statistical analysis and downstream evaluation on code generation task, we demonstrate that Lang2Code is a large-scale and high-quality NL-PL pair corpus and can greatly help developing data-hungry models in future research.


\bibliographystyle{ACM-Reference-Format}
\bibliography{main}

\end{document}